\documentclass[nonacm,sigconf]{acmart}
\usepackage[utf8]{inputenc}
\usepackage{xspace}
\usepackage{graphicx}
\usepackage{color}
\usepackage{graphicx}
\renewcommand{\paragraph}[1]{\vspace*{1mm} \noindent \textbf{#1}}

\title{Towards a Measure of Individual Fairness for Deep Learning}

\author{Krystal Maughan}
\email{Krystal.Maughan@uvm.edu}
\affiliation{%
  \institution{University of Vermont}
}

\author{Joseph P. Near}
\email{jnear@uvm.edu}
\affiliation{%
  \institution{University of Vermont}
}

\begin{abstract}
  Deep learning has produced big advances in artificial intelligence, but trained neural networks often reflect and amplify bias in their training data, and thus produce unfair predictions. We propose a novel measure of individual fairness, called \emph{prediction sensitivity}, that approximates the extent to which a particular prediction is dependent on a protected attribute. We show how to compute prediction sensitivity using standard automatic differentiation capabilities present in modern deep learning frameworks, and present preliminary empirical results suggesting that prediction sensitivity may be effective for measuring bias in individual predictions.
\end{abstract}

\begin{document}

\maketitle

\section{Introduction}

In a recent New York Times article, a facial recognition algorithm incorrectly identified the wrong individual given a blurry image of a shoplifter~\cite{nytimes_wrongfully}. The facial recognition system was found to perform noticeably worse when used to identify non-white demographics.
%It is more difficult for the algorithm to make a distinction between individuals who are non-white versus individuals who are.
A lack of diversity in the data set used to train the system was said to have contributed to the bias in the algorithm. In this particular case, this bias resulted in the wrongful arrest of a man for a crime of which he was innocent.
Today, automated Artificial Intelligence systems are pervasive: they are used to automate the processes of hiring candidates~\cite{hiring1, hiring2}, firing candidates~\cite{firing} and in recommending consumer products based on our predicted emotions and those of our peer groups~\cite{facebook}.
An error or erroneous prediction can therefore lead to the loss of a job, the loss of credit or economic and legal implications that can perpetuate cycles of vulnerability and poverty. On a larger scale, when we look at the total life-cycle of automated artificial intelligence systems, these automated systems will be adopted in some capacity by countries with less economic and political stability. Erroneous individual and group predictions can therefore lead to larger societal issues such as civic unrest and an increased mistrust of governments and larger institutions by citizens of those nations~\cite{chile}.
Unfortunately, numerous examples suggest that automated systems inherit and even amplify the biases already present in society~\cite{automat}.
The goal of \emph{fair AI} is to better understand the origins of these biases and their impacts on society, and to develop mitigations and recommendations to improve the situation.

\paragraph{Defining Fairness.}
Fairness is defined as the degree to which judgments can be considered to discriminate against a particular individual or a group~\cite{mehrabi2019survey}.
%a particular individual or group in a data set should not be regarded with less importance than another.
This acknowledges an ideal in which members of all groups are regarded as equal and no one group has dominance or influence over another.

However, Artificial Intelligence tends to reflect and amplify the unfairness we see in society, resulting in automated systems that demonstrate bias in favor of a particular individual or group. Historical events that perpetuated power dynamics (such as colonialism or imperialism) and disrupted the cultural and social economies of marginalized people have directly shaped today's society.
%to our understanding of social inequity as it exists in the world.
We have quantified progress or proof of advocacy for fairness by measuring incremental improvements on these egregious systems of inequity and the devaluation of marginalized groups. Since data is a proxy for this history, unfairness in Artificial Intelligence systems therefore becomes an inevitable automation of these biases. 

\emph{Measuring fairness} turns out to be extremely challenging.
Numerous formal metrics have been developed to measure fairness in machine learning systems~\cite{dwork2012fairness, hardt2016equality, zafar2015fairness, gajane2017formalizing, verma2018fairness, barocas-hardt-narayanan}, but no single metric completely captures our intuitive notions of fairness. Most metrics measure \emph{group fairness}: they compare the rate of positive outcomes for \emph{privileged} and \emph{unprivileged} groups of individuals. As a result, a model that scores highly on a group fairness metric may still make blatantly unfair predictions for particular individuals as long as outcomes are similar \emph{on average}. Measuring \emph{individual fairness}~\cite{dwork2012fairness} remains a challenging problem.
%A more detailed description of these challenges appears in Section~\ref{sec:chall-defin-fairn}.

\paragraph{Fairness in Deep Learning.}
In this paper, we study the problem of fairness in the setting of deep learning. A number of approaches have been proposed to improve group fairness in deep neural networks~\cite{beutel2017data, shankar2017no, zhang2016understanding, zhang2018mitigating, wadsworth2018achieving, celis2019improved}, many of which yield large improvements in group fairness metrics without significantly affecting accuracy.

\paragraph{Prediction Sensitivity: a measure of individual fairness.}
In this paper, we propose a new approach designed to help understand fairness properties of \emph{individual} predictions.
%that detects when an \emph{individual} prediction by a deep neural network might be unfair.
Our approach measures the extent to which a prediction is based on the value of a \emph{protected attribute} which encodes an individual's membership in either a privileged or unprivileged group.

We call this measurement the \emph{prediction sensitivity}, due to its similarity to the idea of sensitivity in differential privacy~\cite{dwork2006calibrating, dwork2012fairness}.
We describe how to compute prediction sensitivity efficiently in standard deep learning frameworks via a novel application of automatic differentiation.
Finally, we present preliminary empirical results from comparing prediction sensitivity to standard metrics for group fairness, suggesting that prediction sensitivity may be effective for understanding bias in trained models.
%We show that this approach causes standard fairness metrics to improve while producing a minimal effect on accuracy (the accuracy part might end up being false).

Prediction sensitivity is intended as a first step towards understanding \emph{individual fairness} in deep neural networks. Like all formal metrics for fairness, there are vital aspects of societal bias that are \emph{not} captured by the definition. An important example of this for prediction sensitivity is the problem of \emph{redundant encodings}~\cite{dwork2012fairness}: features in the data which are \emph{not} marked as protected attributes, but which are correlated with the protected attribute. Prediction sensitivity measures (only) the effect of the protected attribute on the model's prediction, and ignores any effect that correlated attributes might have. As a result, it may be possible to train models with \emph{low} prediction sensitivity (indicating fair predictions) that nevertheless make \emph{unfair} predictions by relying on redundant encodings. This and other limitations of prediction sensitivity (detailed in Section~\ref{sec:discussion-future}) are important topics for future work.

% \paragraph{Future Work.}
% We think this leads to lots of interesting questions for future work:
%   \begin{itemize}
%   \item If we also use an approach that has been demonstrated to work empirically, can we add less noise to predictions and improve accuracy while retaining fairness?
%   \item If we augment prediction sensitivity with additional measures, can we provide a provable guarantee?
%   \item What happens when other attributes are correlated with the protected attribute? How can we address this?
%   \item How does prediction sensitivity relate to <all the metrics> (in particular, group vs individual fairness - I don't know much about this distinction but I think it's probably important to discuss)
%   \item What's the right metric to target? Would another definition of sensitivity fit this better?
%   \end{itemize}

\section{Challenges of Defining Fairness}
\label{sec:chall-defin-fairn}

The issue of fairness in Artificial Intelligence models is particularly complex because of the manner in which we create these models. Models are typically trained and learn from data but the process is a black box which is inscrutable and difficult to reproduce. There have been methods which attempt to address the inscrutability of models by measuring causality via identification of potential outcomes. One method used for inferring the causal reasoning of data is the potential outcomes framework or the Rubin causal model (Johansson et al., 2016)~\cite{gentzel2019case}. This method identifies the possibilities of potential outcomes of a model by looking at individual features and all possible outcomes. Causal inference makes unit-level causal effects difficult to observe, giving rise to the fundamental problem of causal inference. Feature level analysis also has trade-offs. Whenever features are individually assessed, phenomena such as Simpson's Paradox can become problematic and obscure issues which are attributable to bias~\cite{hernan2019second}.
In understanding the complexity of black box models, we have made efforts to simplify or decomplexify models to reason about why they make certain predictions.

\paragraph{Types of Bias.}
Bias can be defined as a systematic error or judgment. Bias often manifests based on assumptions that are proven to be untrue. These issues of bias can take place due to the data collection process, the data codification process, the feature selection and modelling process and the model prediction process. 
Human beings have inherent biases. These biases are aggregated and embedded within our data collection and analysis processes. One example of human bias is group attribution bias, whereby we judge a person as part of a monolithic group rather than as an individual,and project those tendencies of the generalized group to all individuals within that group. 
Implicit biases are an anecdotal extension of our experiences onto the larger group. These experiences are encoded into the data that we collect, the features we use, and the algorithmic predictions of our data. 
An example of bias that manifests in the data collection process is selection bias, whereby the data collected is not representative of the real world. An example of selection bias includes ethical issues in data collection that involve remuneration for giving access to one's data. In scenarios where remuneration is given, care must be taken so that the most economically disadvantaged persons are not giving away their data solely for the promise of remuneration rather than because they choose to do so. 
Participation bias (also called non-responsive bias) is another issue with data collection. This means that there are gaps in the data because of non-participation by certain groups of people. An example of participation bias might be a lack of persons willing to take part in a survey because the specificity of details in the survey may encourage doxing and retaliation by another group.
Another bias is that of seeing cause in correlation or spuriousness, where a feature is identified as associated with a variable when there is no intrinsic correlation. This bias affects the model prediction process, resulting in wrongful predictions.

\paragraph{Group Fairness versus Individual Fairness.}
Fairness is often identified by one's relationship to a particular group, where the assumption is that no specific individual is regarded as more unequal than another within the group. This is typically referred to as Equality.
More broadly speaking, everyone is treated the same regardless of the specificity of their characteristics; that is, in spite of these characteristics. Allocation is therefore based on the needs of the group rather than the needs of a specific individual.
In another definition, each individual's characteristics are individually assessed, and fairness is assessed by these characteristics. Each individual is judged fairly based on their specific needs and requirements, and allocations take the needs of the individual into account. This is typically called Equity.

\paragraph{Measuring Fairness.}
A number of formal fairness metrics have been developed to measure fairness~\cite{dwork2012fairness, hardt2016equality, zafar2015fairness, gajane2017formalizing, verma2018fairness}; Barocas et al.~\cite{barocas-hardt-narayanan} provide a comprehensive overview. These metrics are typically used to measure fairness by comparing the rates of positive predictions for members of the \emph{privileged} and \emph{unprivileged} group. They are typically defined in terms of a \emph{protected attribute}, a feature in the data that indicates each individual's membership status in either the privileged or unprivileged group.
Two specific metrics we will use in this paper are statistical parity~\cite{dwork2012fairness} (also called demographic parity) and disparate impact~\cite{feldman2015certifying}. Both are measures of group fairness.%\jn{need to verify this}

\section{Background}

\paragraph{Deep Learning.}
In this paper, we focus on machine learning models represented by \emph{artificial neural networks}~\cite{goodfellow2016deep}. A model $\mathcal{F}$ is parameterized by a set of \emph{weights} $\theta$ which are optimized during training; we write $\mathcal{F}(\theta, x)$ to represent a \emph{prediction} made by the trained model on an example $x$. Deep learning models are typically trained by optimizing a \emph{loss function} $\mathcal{L}$ using a ground-truth label $y$ for each example. We write the loss for an example $x,y$ as $\mathcal{L}(\mathcal{F}(\theta, x), y)$. During training, the loss is used to update the weights $\theta$ so that loss is reduced during the next training epoch.

\paragraph{Automatic Differentiation.}
\emph{Automatic differentiation}~\cite{baydin2017automatic} is a computational method used to evaluate the derivative of a function efficiently. Automatic differentiation is normally used for computing the \emph{gradient} of the loss with respect to the model's weights:
\[ \nabla_\theta(\mathcal{L}(\mathcal{F}(\theta, x), y)) \]
\noindent This gradient is a vector containing the partial derivative of the model's output with respect to each of the weights. Automatic differentiation systems in modern deep learning frameworks are specifically designed to efficiently compute gradients for functions with many inputs, and they are usually used to calculate gradients during training.

\paragraph{Fairness in Machine Learning.}
The bulk of previous work on fairness in machine learning attempts to improve \emph{group fairness} at training time, often by the introduction of new kinds of regularization~\cite{calders2009building, woodworth2017learning, zafar2015fairness, zafar2017fairness, agarwal2018reductions, russell2017worlds, celis2019classification, beutel2017data, shankar2017no, zhang2018mitigating, wadsworth2018achieving, celis2019improved, zemel2013learning, louizos2015variational, lum2016statistical, adler2018auditing, calmon2017optimized, feldman2015computational, hardt2016equality}. Many of these approaches are suitable for deep learning, and have been empirically validated using the metrics described above.

Existing approaches focus on notions of group fairness, and are validated using metrics for group fairness. As a result, they can sometimes produce models that give blatantly \emph{unfair} predictions for specific individuals, even though they score well on group fairness metrics.

\section{Prediction Sensitivity}

We propose \emph{prediction sensitivity}, which quantifies the \emph{extent to which a prediction depends on the protected attribute}. We hypothesize that models which rely heavily on the value of the protected attribute are likely to make different predictions for members and non-members of the advantaged group. Prediction sensitivity may therefore be a useful measure of \emph{individual fairness} for each prediction made by the model.

The name ``prediction sensitivity'' comes from the connection to the notion of \emph{function sensitivity} used in differential privacy~\cite{dwork2006calibrating}. Prediction sensitivity can be viewed as an approximation of function sensitivity for \emph{neighboring inputs} that differ in their protected attribute. However, prediction sensitivity is \emph{not} an upper bound on function sensitivity, as discussed later in this section. Prediction sensitivity is also related to the individual metric proposed by Dwork et al.~\cite{dwork2012fairness}, but prediction sensitivity can be efficiently computed for artificial neural networks.

\paragraph{Formal Definition.}
Formally, we assume the existence of a neural network architecture $\mathcal{F}$ such that for a vector of trained weights $\theta$ and feature vector $x$, we can make a prediction $\hat{y}$ as follows:
\[\hat{y} = \mathcal{F}(\theta, x)\]

For such a model $\mathcal{F}$, we define the prediction sensitivity with respect to attribute $a \in x$ as the partial derivative:
\[ \Big\lvert \frac{\partial}{\partial a} \mathcal{F}(\theta, x) \Big\rvert \]

\paragraph{Computing Prediction Sensitivity.}
Automatic differentiation libraries are commonly used to compute gradients of the loss during training; the same libraries can be used to efficiently compute prediction sensitivity. Given a loss function $\mathcal{L}$ and a training example $x, y$, for current weights $\theta$, the training process might compute the gradient:
\[ \nabla_\theta(\mathcal{L}(\mathcal{F}(\theta, x), y)) \]
\noindent Here, we write $\nabla_\theta$ to denote the gradient with respect to each weight in $\theta$. Thus the gradient contains partial derivatives of the \emph{loss} with respect to the \emph{weights}. Deep learning frameworks like TensorFlow are designed to compute gradients like these via automatic differentiation, but they are also capable of computing other types of gradients.

To compute prediction sensitivity for a feature vector $x$, we want to obtain the partial derivative of the \emph{prediction} with respect to \emph{one feature}. Given a trained model consisting of $\mathcal{F}$ and $\theta$, we can compute prediction sensitivity for the feature $a$ as:
\[ \lvert \nabla_a (\mathcal{F}(\theta, x)) \rvert \]

This value can be efficiently computed using existing deep learning frameworks. We have implemented this approach in TensorFlow, and used it to obtain the empirical results in Section~\ref{sec:empirical-evaluation}.

\paragraph{Properties of Prediction Sensitivity.}
Prediction sensitivity intends to capture the extent to which predictions depend on the protected status of an individual, but it is not always a perfect measurement of this dependence, and care must be taken in applying it. First, a model may predict positive outcomes for the advantaged group at a higher rate \emph{without} relying on the protected attribute, and prediction sensitivity will not be useful in this case. Section~\ref{sec:empirical-evaluation} contains our preliminary empirical results suggesting that biased models often \emph{do} learn to rely on the protected attribute in making predictions.

Second, prediction sensitivity measures sensitivity to changes in the protected attribute at \emph{one particular point}, and small changes to one or more features could potentially increase or decrease prediction sensitivity significantly (in fact, the amount of this change is potentially unbounded). In most applications, we must assume that prediction sensitivity is fairly \emph{smooth}---i.e. that small changes in features do not produce large changes in prediction sensitivity---in order for the measure to be useful. We hypothesize that most neural networks result in reasonably smooth prediction sensitivities; we discuss future work on this topic in Section~\ref{sec:discussion-future}.

Third, the scale of prediction sensitivity depends on the model's architecture and the trained weights, in addition to the features of the input example. As a result, it is not possible to compare prediction sensitivities across trained models. A prediction sensitivity of 0.1 may be relatively low for one model, and indicate a prediction that is minimally dependent on the protected attribute (e.g. the prediction sensitivity for the other features may be 10.0); the same prediction sensitivity may be very high for another model. We discuss the impact of this fact in Section~\ref{sec:discussion-future}.

\section{Applying Prediction Sensitivity}
\label{sec:apply-pred-sens}

Prediction sensitivity is designed to be a measure of individual fairness, but there are several different ways it could be applied to improve fairness in AI systems.

\paragraph{Fairness Monitor.}
One obvious opportunity for applying prediction sensitivity is as a way to measure the fairness of individual predictions made by a deployed model, and to signal an alarm for predictions that are \emph{not} fair. A fairness monitor can be implemented by deciding on a \emph{prediction sensitivity threshold}, calculating the prediction sensitivity for each prediction made, and signaling an alarm for each prediction whose sensitivity is above the threshold. We present a preliminary empirical evaluation of the potential for this approach to improve the fairness of predictions in Section~\ref{sec:empirical-evaluation}.

The fairness monitor idea also has several unresolved challenges. First, it cannot \emph{improve} predictions---only raise an alarm for bad ones---and it is still unclear how the feedback given by such an alarm could be used to improve the model. Second, it may be challenging to decide on a threshold in practice, especially because we expect that the appropriate setting for the threshold will depend on the training data, model architecture, and training algorithm used. Third, it may be difficult in practice to validate that a particular choice of threshold is the right one---the fairness monitor may result in both false negatives and false positives, and both classes of error may be difficult to identify.

\paragraph{Tracking Statistics.}
When a useful threshold is difficult to identify, prediction sensitivity could still be useful in measuring statistics about individual fairness of the predictions made by the deployed model. For example, the distribution of prediction sensitivities over many predictions might reveal low sensitivity for most predictions, with a few outliers---suggesting that these outliers should be investigated with the goal of improving the model. It might also reveal changes over time as the data used in making predictions starts to differ from the training data used to build the model.

\paragraph{Evaluating Models.}
We hypothesize that prediction sensitivity might also be useful for evaluating a trained model empirically to understand how it makes predictions. For example, the distribution of prediction sensitivities over a testing set might represent a useful measure for fairness comparable to metrics like statistical parity and disparate impact. Unlike those metrics, however, prediction sensitivity is a measure of \emph{individual} fairness, and may therefore reveal additional information that metrics for group fairness cannot (e.g. for models that usually make fair predictions, but are blatantly unfair for some outliers). Our preliminary empirical results in Section~\ref{sec:empirical-evaluation} provide some support for this idea.

\section{Empirical Evaluation}
\label{sec:empirical-evaluation}

To explore the feasibility of using prediction sensitivity to understand the fairness of a model's predictions, we performed an empirical evaluation using a simple neural network trained on the Adult dataset~\cite{adult}. The bias in this dataset is well-documented~\cite{bellamy2018ai}: models trained without mitigations demonstrate bias against unprivileged groups on both gender and race.

Using TensorFlow, We trained a simple neural network consisting of two fully-connected hidden layers to perform the binary classification task of the Adult dataset. We implemented two commonly-used metrics for group fairness: statistical parity~\cite{dwork2012fairness} and disparate impact~\cite{feldman2015certifying}. When evaluated on the \emph{gender} attribute, our model (trained without any mitigations for fairness) yielded a statistical parity value of -0.04 and a disparate impact value of 0.53. Both of these values indicate bias against the unprivileged group.

\begin{figure*}
  \centering
  \includegraphics[width=.95\textwidth]{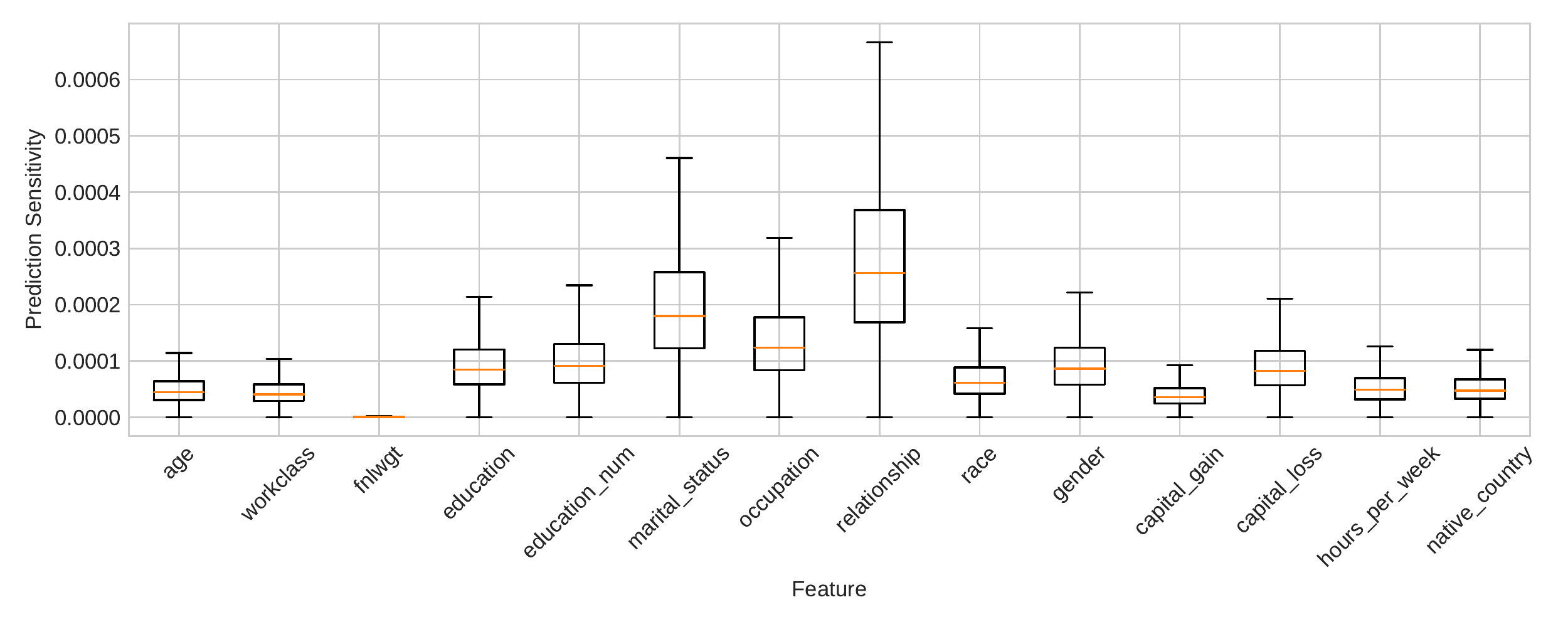}
  \vspace*{-3mm}
  \caption{Distribution of prediction sensitivities for each feature of the Adult dataset.}
  \label{fig:prediction_sensitivities}
\end{figure*}

\paragraph{Distribution of Prediction Sensitivity over Features.}
Figure~\ref{fig:prediction_sensitivities} summarizes, using box-and-whisker plots, the distribution of prediction sensitivities from the model we trained for 14 features in the Adult dataset's test set. Features like education and occupation, which we might expect to have a large influence on income, indeed tend to have high prediction sensitivity. The gender and race features, which indicate an individual's membership in privileged or unprivileged groups, also have significant prediction sensitivity---contributing to the model's bias. Surprisingly, the relationship feature has the \emph{highest} average prediction sensitivity---perhaps because this feature acts as a \emph{proxy} for gender. This is an important issue for future work, discussed more in Section~\ref{sec:discussion-future}.

\begin{figure}
  \centering
  \includegraphics[width=.48\textwidth]{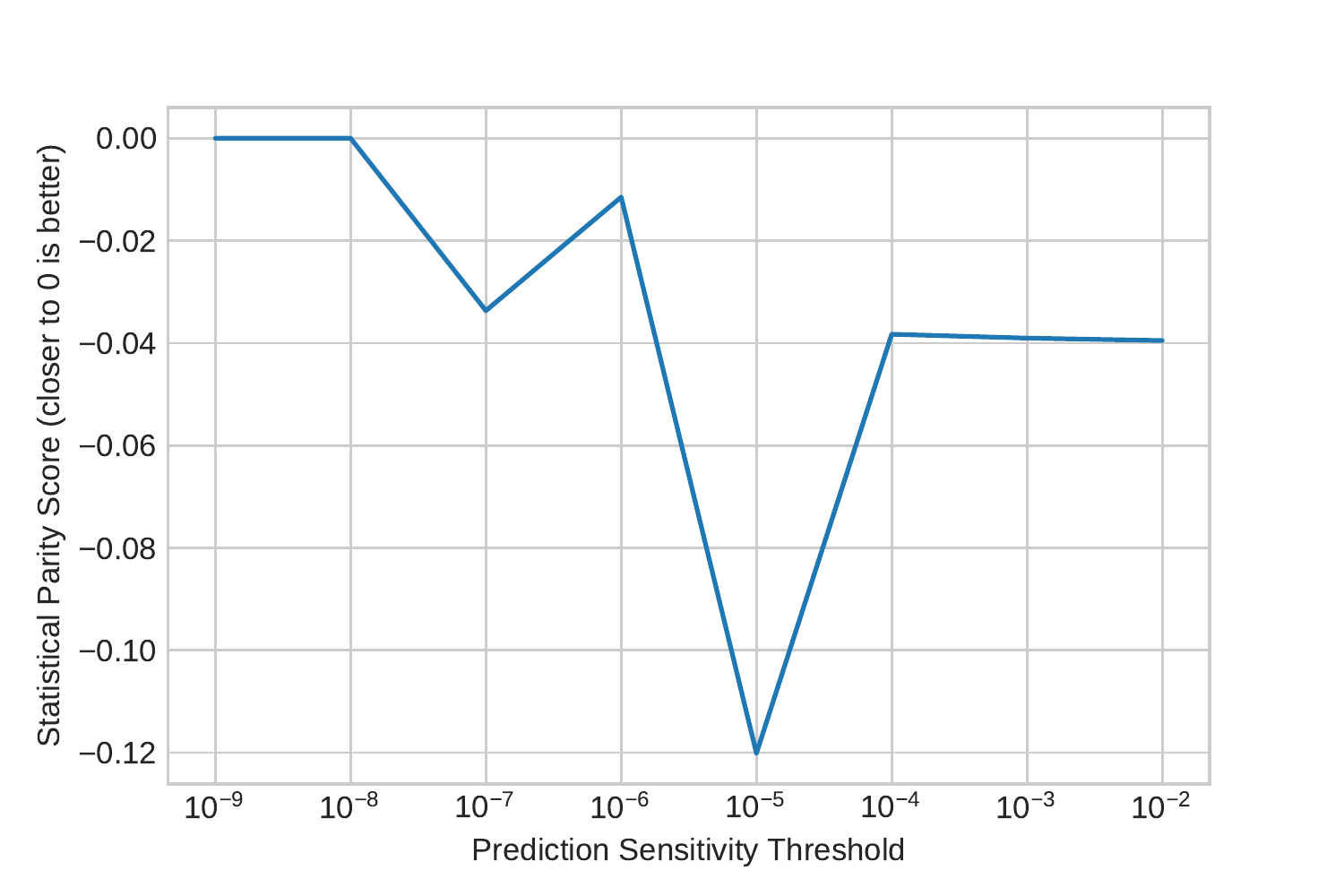}
  \caption{Effect of the fairness monitor on \emph{statistical parity} for various settings of the prediction sensitivity threshold in the Adult dataset.}
  \label{fig:statistical_parity}
\end{figure}

\begin{figure}
  \centering
  \includegraphics[width=.48\textwidth]{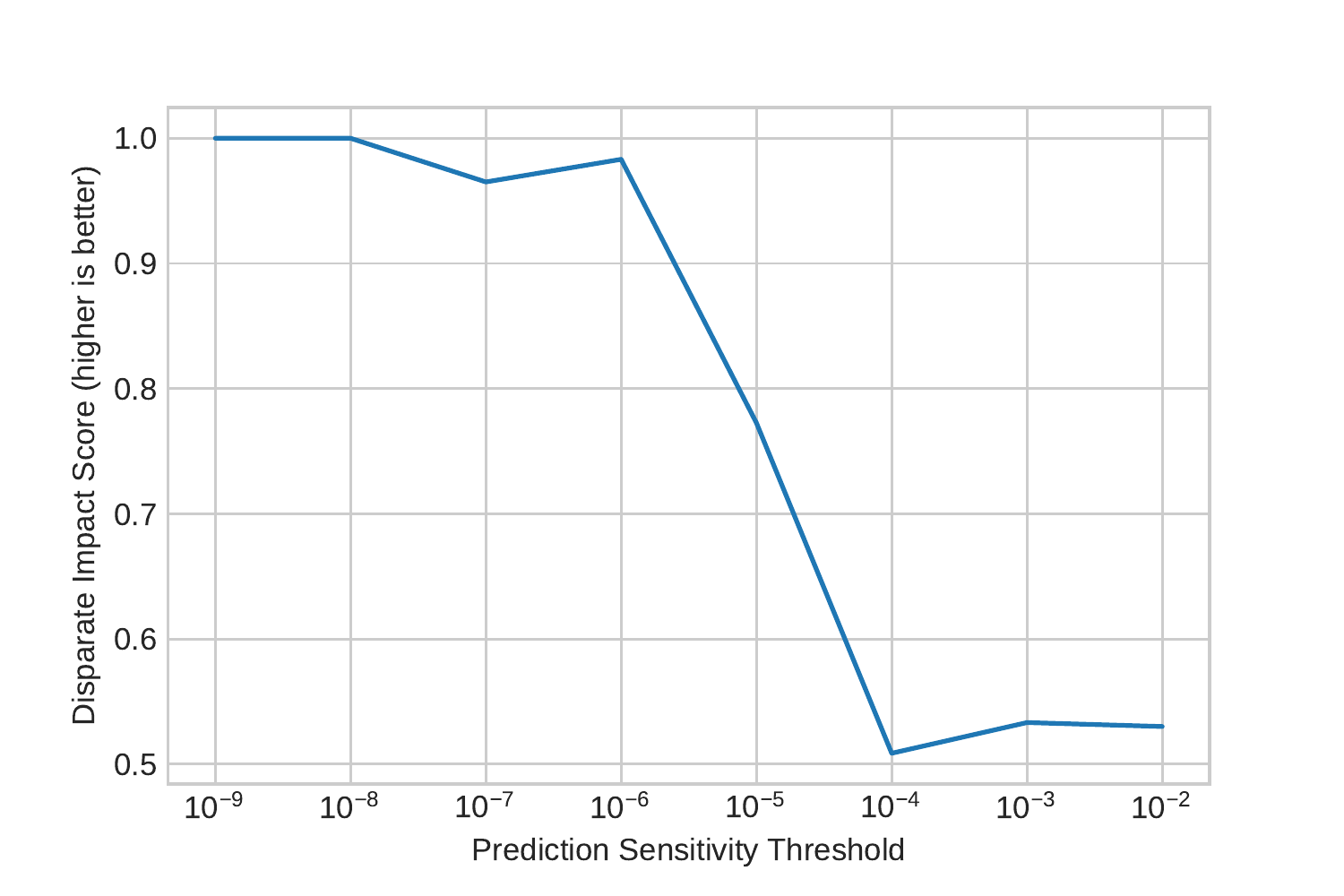}
  \caption{Effect of the fairness monitor on \emph{disparate impact} for various settings of the prediction sensitivity threshold in the Adult dataset.}
  \label{fig:disparate_impact}
\end{figure}

\paragraph{Effect of the Fairness Monitor.}
To measure how effective prediction sensitivity would be in building the fairness monitor described in Section~\ref{sec:apply-pred-sens}, we evaluated this approach on the test set of the Adult dataset using two metrics for group fairness. For each example in the test set, we used our trained model to make a prediction, and measured the prediction sensitivity with respect to the ``gender'' feature. If the prediction sensitivity exceeded a threshold, we discarded the prediction (as a proxy for sounding the alarm described in Section~\ref{sec:apply-pred-sens}); if not, we kept the prediction. Then, we evaluated \emph{only} the ``kept'' predictions using a metric for group fairness (statistical parity or disparate impact). We repeated this experiment for various values of the threshold to determine the effect of the threshold's value on the group fairness metrics.

The results appear in Figures~\ref{fig:statistical_parity} (for statistical parity) and~\ref{fig:disparate_impact} (for disparate impact). In both cases, we see that the group fairness metric improves as the threshold for prediction sensitivity goes down. This effect shows that the \emph{individual} measure of fairness provided by prediction sensitivity may be well-aligned with the \emph{group-level} measures of fairness already in common use, suggesting that prediction sensitivity may indeed be useful in measuring the fairness of individual predictions made by a model.

\section{Discussion \& Future Work}
\label{sec:discussion-future}

\paragraph{Limitations of Preliminary Results.}
Our preliminary results suggest that prediction sensitivity may be a useful measure of individual fairness, but more study is needed to understand it fully. In particular, it is challenging even to define the meaning of false negatives and false positives in the setting of a fairness monitor---doing so would require a ``ground truth'' about the fairness of a particular prediction, which is impossible to formalize. Our experiments also do not measure the affect of the fairness monitor on accuracy, since discarding predictions is not realistic during deployment anyway. We expect that in any intervention based on prediction sensitivity, a tradeoff will exist between fairness and accuracy, and we plan to investigate this further in the future. Finally, our experiments examined only a single dataset and trained model; we plan to extend our evaluation to additional datasets and a variety of models in future work.

\paragraph{Smoothness of Prediction Sensitivity.}
As mentioned earlier, prediction sensitivity measures the effect of changes in the protected attribute on the model's prediction \emph{at a particular point}, but does not say anything about prediction sensitivity at \emph{nearby points}. For example, prediction sensitivity may be very low for a particular example, but a slightly modified example may result in very high prediction sensitivity. As a result, it may be impossible to set a reasonable threshold for an application like a fairness monitor that avoids blatant false negatives; this effect also means that it is not possible to compare prediction sensitivity values across trained models. In future work, we plan to develop an automated way to analyze the smoothness of prediction sensitivity for a particular trained model.

\paragraph{Correlated Attributes.}
Prediction sensitivity does not address the possibility of non-protected attributes that are correlated with the protected attribute (also called a \emph{redundant encoding} by Dwork et al.~\cite{dwork2012fairness}). The model may use such an attribute to make its prediction, and ignore the protected attribute; in this case, the model may result in unfair \emph{outcomes} for the unprivileged group, even when its prediction sensitivity with respect to the protected attribute is low. This possibility is a direct result of the way prediction sensitivity is defined: it measures \emph{one way} the model could arrive at an unfair prediction (i.e. by relying on the value of the protected attribute), but ignores other ways that the same outcome could occur (e.g. by relying on a correlated attribute).

Our empirical results suggest that models which make unfair predictions (as measured by group fairness metrics like disparate impact) tend to rely on the protected attribute to do so. However, these results are for a single model trained on a single dataset; further experiments on additional models and datasets are needed. In addition, interventions during training to mitigate bias may have a large effect on the connection between group fairness and prediction sensitivity; we plan to study these as well.

\section{Related Work}
%\paragraph{Methods of Fairness using Differential Privacy Techniques}

As the problem of bias in machine learning becomes more apparent, an increasing amount of research has been devoted to the topic. Mehrabi et al.~\cite{mehrabi2019survey} provide a survey, including broad definitions of bias and fairness use in the field of Artificial Intelligence, and how these definitions lay the groundwork for empirical measures of fairness in the field today. The book by Barocas et al.~\cite{barocas-hardt-narayanan} gives an excellent overview of the field of algorithmic fairness more generally.

The bulk of previous work on fairness in machine learning attempts to improve fairness at training time, often by the introduction of new kinds of regularization~\cite{calders2009building, woodworth2017learning, zafar2015fairness, zafar2017fairness, agarwal2018reductions, russell2017worlds, celis2019classification}. Recent work in this area for deep learning (much of it leveraging adversarial learning) has demonstrated impressive empirical results, showing that the tradeoff between fairness and accuracy can often be successfully navigated in practice~\cite{beutel2017data, shankar2017no, zhang2018mitigating, wadsworth2018achieving, celis2019improved}. Other approaches are based on pre-processing, and do not require changes to the training process~\cite{zemel2013learning, louizos2015variational, lum2016statistical, adler2018auditing, calmon2017optimized}. These approaches are suitable for deep learning, but typically do not provide the same level of accuracy as modifications to the training process. Finally, several approaches based on post-processing have been proposed~\cite{feldman2015computational, hardt2016equality}.

Extensive work has also explored how to measure fairness, and why obtaining fairness in Artificial Intelligence is difficult. A number of different metrics have been proposed~\cite{dwork2012fairness, hardt2016equality, zafar2015fairness, gajane2017formalizing, verma2018fairness, barocas-hardt-narayanan}, most of them designed to ensure group fairness. Other metrics apply to specific cases; for example, Mansoury et al.~\cite{mansoury2020fairmatch} describe the importance of \emph{aggregate diversity} and propose a methods for improving aggregate diversity in recommender systems. Individual fairness has also been studied~\cite{dwork2012fairness}, and has been extended to approaches for cross-cultural fairness that identifies and measures fairness based on subgroups~\cite{kearns2019empirical}. Zhang et al.~\cite{zhang2016understanding} shows how current methods of generalizing neural networks are faulty, and suggests a link between generalization and fairness.

\section{Conclusion}
% We have identified challenges in measuring formally defined methods of fairness in Deep Learning, and the needs for provable constraints for measuring Group Bias based on ...

We have proposed prediction sensitivity, a measure of individual fairness in deep neural networks that may help us understand sources of bias in automated systems based on Artificial Intelligence. We showed how to compute prediction sensitivity efficiently using existing deep learning frameworks, and we presented preliminary empirical results suggesting that prediction sensitivity may be a useful metric in measuring bias. We believe that prediction sensitivity is an important first step towards better understanding individual fairness, and we have identified a number of important areas for future study in that direction.

\section*{Acknowledgements}

We thank David Darais and Kristin Mills for their contributions to the development of this work, and the Mechanism Design for Social Good reviewers for their helpful comments. This research was supported in part by an Amazon Research Award.

\bibliographystyle{plain}
\bibliography{refs, fairness}

\end{document}